\documentclass[runningheads]{llncs}
\usepackage{color}
\usepackage{graphicx}
\usepackage{multirow,makecell}
\usepackage{subfigure}
\usepackage{amsmath}
\usepackage{hyperref}
\usepackage[misc,geometry]{ifsym}

\renewcommand\UrlFont{\color{blue}\rmfamily}
\usepackage{booktabs}
\usepackage{color}
\usepackage{multirow}

\begin{document}
\title{SynthTIGER: Synthetic Text Image GEneratoR \\ Towards Better Text Recognition Models}
\titlerunning{SynthTIGER}
\author{Moonbin Yim\inst{1}\orcidID{0000-0002-7272-2198} \and Yoonsik Kim\inst{1}\orcidID{0000-0001-8023-8278} \and Han-Cheol Cho\inst{1}\orcidID{0000-0002-8891-0177} \and \\ Sungrae Park\inst{2}(\Letter)\orcidID{0000-0001-5338-0113}}
\authorrunning{Yim et al.}
\institute{CLOVA AI Research, NAVER Corporation \\
\email{\{moonbin.yim, yoonsik.kim90, han-cheol.cho\}@navercorp.com} \and
Upstage AI Research \\
\email{sungrae.park@upstage.ai}}
\maketitle              %

\begin{abstract}

For successful scene text recognition (STR) models, synthetic text image generators have alleviated the lack of annotated text images from the real world. Specifically, they generate multiple text images with diverse backgrounds, font styles, and text shapes and enable STR models to learn visual patterns that might not be accessible from manually annotated data. 
In this paper, we introduce a new synthetic text image generator, SynthTIGER, by analyzing techniques used for text image synthesis and integrating effective ones under a single algorithm.
Moreover, we propose two techniques that alleviate the long-tail problem in length and character distributions of training data. In our experiments, SynthTIGER achieves better STR performance than the combination of synthetic datasets, MJSynth (MJ) and SynthText (ST). Our ablation study demonstrates the benefits of using sub-components of SynthTIGER and the guideline on generating synthetic text images for STR models. Our implementation is publicly available at \href{https://github.com/clovaai/synthtiger}{\UrlFont https://github.com/clovaai/synthtiger}. 

\keywords{Optical Character Recognition \and Synthetic Text Image Generator \and Scene Text Synthesis \and Scene Text Recognition \and Synthetic Dataset}

\end{abstract}

\section{Introduction}

\begin{figure}[t]
\centering
\begin{tabular}{cc}
MJ~\cite{jaderberg2014synthetic} & \makecell{\includegraphics[width=0.8\linewidth]{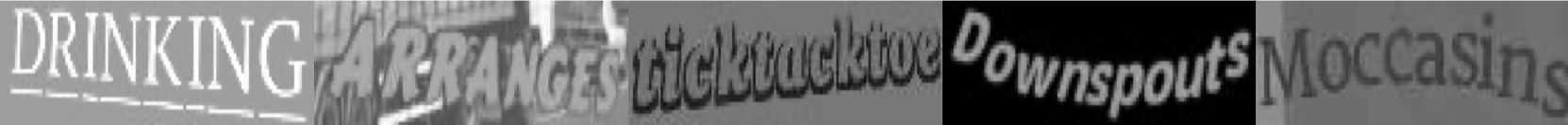}} \\
ST~\cite{gupta2016synthetic}& \makecell{\includegraphics[width=0.8\linewidth]{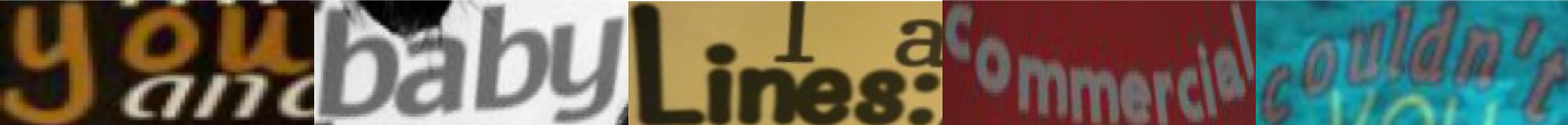}} \\
Ours& \makecell{\includegraphics[width=0.8\linewidth]{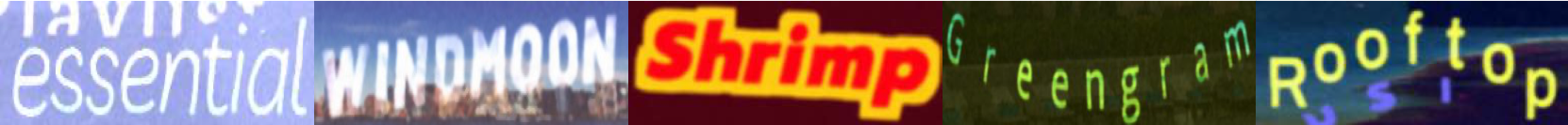}} \\
\end{tabular}
\caption{Word box images generated by synthesis engines. MJ provides diverse text styles but there is no noise from other texts. The examples of ST are cropped from a scene text image including multiple text boxes and they includes some part of other texts. Although our synthesis engine generates word box images as like MJ, its examples includes text noises observed in examples of ST.}
\label{fig:teaser-examples}
\end{figure}

Optical character recognition (OCR) is a technology extracting machine-encoded texts from text images. 
It is a fundamental function for visual understanding and has been used in diverse real-world applications such as automatic number plate recognition~\cite{automaticPlate}, business document recognition~\cite{motahari2021report,hwang2020spatial,hwang2019post} and passport recognition~\cite{limonova2017slant}. 
In the deep learning era~\cite{liu2019scene,long2021scene}, OCR performance has been dramatically improved by learning from large-scale data consisting of image and text pairs. 
In general, OCR uses large-scale data consisting of synthetic text images because it is virtually impossible to manually gather and annotate real text images that cover the exponential combinations of diverse characteristics such as text length, fonts, and backgrounds.

OCR in the wild consists of two sub-tasks, scene text detection (STD) and scene text recognition (STR).
They require similar but different training data.
Since STD has to localize text areas from backgrounds, its training example is a raw scene or document snapshot containing multiple texts. In contrast, STR identifies a character sequence from a word box image patch that contains a single word or a line of words. It requires a number of synthetic examples to cover the diversity of styles and texts that might exist in the real world. This paper focuses on synthetic data generation for STR to address the diversity of textual appearance in a word box image. 

There are two popular synthesis engines, MJ~\cite{jaderberg2014synthetic} and ST~\cite{gupta2016synthetic}.
MJ is a text image synthesis engine that generates word box images by processing multiple rendering modules such as font rendering, border/shadow rendering, base coloring, projective distortion, natural data blending, and noise injection.
By focusing on generating word box images rather than scene text images, MJ can control all text styles, such as font color and size, used in its rendering modules but the generated word box images cannot fully represent text regions cropped from a real scene image.
In contrast, ST~\cite{gupta2016synthetic} generates scene text images that includes multiple word boxes on a single scene image. ST identifies text regions and writes texts upon the regions by processing font rendering, border/shadow rendering, base coloring, and poisson image editing. Since word boxes are cropped from a scene text image, the identified word box images can include text noises from other word boxes as like real STR examples. 
However, there are some constraints on choosing text styles because background regions identified for text rendering may not be compatible with some text rending functions (e.g., too small to use big font size).

To take advantage of both approaches, recent STR research~\cite{shi2018aster,baek2019STRcomparisons} simply integrates datasets generated by both MJ and ST. However, the simple data integration not only increases the total number of training data but also causes a bias on co-covered data distributions of both synthesis engines. Although the integration provides better STR performance than the individuals, there is still room for improvement by considering a better method combining the benefits of MJ and ST. 

In this paper, we introduces a new synthesis engine, referred to as \textbf{Synth}etic \textbf{T}ext \textbf{I}mage \textbf{GE}nerato\textbf{R} (SynthTIGER), for better STR models. As like MJ, SynthTIGER generates word box images without the style constraints of ST. However, as like ST, it adopts additional noise from other text regions, which can occur during cropping a text region from a scene image. Fig.~\ref{fig:teaser-examples} shows synthesized examples with MJ, ST, and SynthTIGER. The examples of SynthTIGER may contain parts of other texts as like those of ST. In our apples-to-apples comparison, using SynthTIGER shows better performance than the cases of using MJ and ST respectively. Moreover, SynthTIGER provides comparable performance to the integrated dataset of MJ and ST even though only one synthesis engine is utilized. 

Furthermore, we propose two methods to alleviate skewed data distribution for infrequent characters and short/lengthy words. Previous synthesis engines generate text images by randomly sampling target texts from a pre-defined lexicon. Due to the low sampling chance of infrequent characters and very short/long words, trained models often poorly perform on these kinds of words. SynthTIGER uses length augmentation and infrequent character augmentation methods to address this problem. As shown in experiment results, these methods improve STR performance over rare and short/long words.

Finally, this paper provides an open-source synthesis engine and a new synthetic dataset that shows better STR performance than the combined dataset of MJ and ST. Our experiments under fair comparisons to baseline engines prove the superiority of SynthTIGER. Furthermore, ablative studies on rendering functions describe how rendering processes in SynthTIGER contribute to improving STR performance. The experiments on data distributions of lengths and characters show the importance of synthetic text balancing. The official implementation of SynthTIGER is open-source and the synthesized dataset is publicly available.

\section{Related Work}

In STR, it has become a standard practice to use synthetic datasets.
We introduce previous studies providing synthetic data generation algorithms for STR~\cite{jaderberg2014synthetic} and other tasks that can be exploited for STR~\cite{gupta2016synthetic,liao2020synthtext3d,long2020unrealtext}.

MJ~\cite{jaderberg2014synthetic}  is one of the most popular data generation algorithms (and the dataset generated with that approach) for STR. It produces an image patch containing a single word. In detail, the algorithm consists of six stages. \textit{Font rendering} stage randomly selects a font and font properties such as size, weight and underline. Then, it samples a word from a pre-defined vocabulary and renders it on the foreground image layer following a horizontal line or a random curve. \textit{Border/shadow rendering} step optionally adds inset border, outset border, or shadow image layers with a random width. The following \textit{base coloring} stage fills these image layers with different colors. \textit{Projective distortion} stage applies a random, full-projective transformation to simulate the 3D world. \textit{Natural data blending} step blends these image layers with a randomly sampled crop of an image from the ICDAR 2003 and SVT datasets. Finally, \textit{Noise} stage injects various noise such as blur and JPEG compression artifacts into the image.
While MJ is known to generate text images useful enough to train STR models, it is not clear how much each stage contributes to its success.

Synthetic datasets for scene text detection (STD) can be used for STR by cropping text regions from synthesized images. The main difference of STD data generation algorithms from STR is that STD must consider the geometry of backgrounds to create realistic images.
ST~\cite{gupta2016synthetic} is the most successful STD data generation algorithm. In detail, it first samples a text and a background image. Then, it segments the image based on color and texture to obtain well-defined regions (e.g., surfaces of objects). Next, it selects the region for the text and fills the text (and optionally outline) with a color based on the region's color. Finally, the text is rendered using a randomly selected font and transformation according to the region's orientation.
However, the use of off-the-shelf segmentation techniques for text-background alignment can produce erroneous predictions and result in unrealistic text images.
Recent studies like SynthText3D~\cite{liao2020synthtext3d} and UnrealText~\cite{long2020unrealtext} address this problem by synthesizing images with 3D graphic engines. Experiment results show that text detection performance can be notably improved by using synthesized text images without text alignment error.
However, it is not clear whether these datasets generated from the virtual 3D world can benefit text recognition task.

\section{SynthTIGER}

SynthTIGER consists of two major components: text selection and text rendering modules. The text selection module is used to sample a target text, $\mathbf{t}$, from a pre-defined lexicon, $\mathbf{L}$. Then, the text rendering module generates a text image by using multiple fonts $\mathbf{F}$, backgrounds (textures) $\mathbf{B}$, and a color map $\mathbf{C}$.
In this section, we first describe the text rendering process and then the target text selection process.

\begin{figure}[t]
\centering
\includegraphics[width=1.0\linewidth]{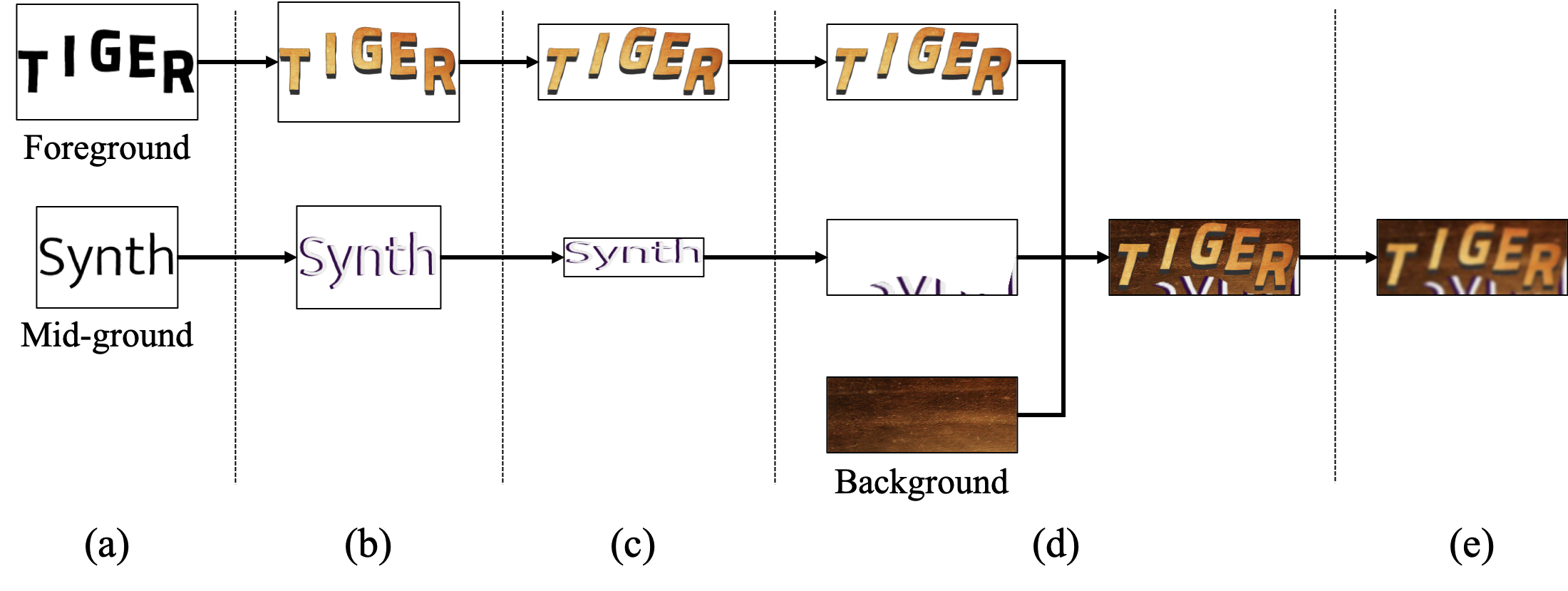}
\caption{Overview of SynthTIGER rendering processes consisting of (a) text shape selection, (b) text style selection, (c) transformation, (d) blending, and (e) post-processing. Foreground text presents a ground-truth text of a generated image and mid-ground text is used as a high frequency visual noises with lines and dots from visual appearance of characters. Foreground and mid-ground texts are rendered independently in (a), (b), and (c) and the background and the texts are combined in (d). Finally, vision noises are added upon the combined image in (e).}
\label{fig:rendering-process}
\end{figure}

\subsection{Text Rendering Process}

Synthesized text images should reflect realness of texts in both a ``micro'' perspective of a word box image and a ``macro'' perspective of a scene-level text image. The rendering process of SynthTIGER generates text-focused images for the micro-level perspective, but it additionally adapts noises of the macro-level perspective. Specifically, SynthTIGER renders a target text and a noisy text and combines them to reflect the realness of the text regions (in a wild, a part of a word appearance can be included in a region for another word). Fig.~\ref{fig:rendering-process} overviews the modules of SynthTIGER engine. It consists of five procedures: (a) text shape selection, (b) text style selection, (c) transformation, (d) blending, and (e) post-processing. The first three processes, (a), (b), and (c), are separately applied to the foreground layer for a target text and the mid-ground layer for a noise text. In the (d) step, the two layers are combined with a background to represent a single synthesized image. Finally, the (e) adds realistic noises. The followings introduce each module in detail.

\subsubsection{(a) Text Shape Selection}

Text shape selection decides a 2-dimensional shape of a 1-dimensional character sequence. This process first identify individual character shapes of a target text $\mathbf{t}$ and then renders them upon a certain line on 2D space in the left-to-right order. 

To reveal visual appearances of the characters, a font is randomly selected from a pool of font styles $\mathbf{F}$ and each character is rendered upon individual boards with randomly chosen font size and thickness. To add diversity of font styles, elastic distortion~\cite{simard2003best} is applied to the rendered characters. 

Defining a spatial order of characters is essential to map characters upon 2-dimensional space. For straight texts, SynthTIGER basically aligns character boards in the left-to-right order with a certain margin between the boards. For curved texts, SynthTIGER places the character boards on a parabolic curve. The curvature of the curve is identified by the maximum height-directional gaps between the centers of the boards. The maximum gap is randomly chosen and the middle points of the target text are allocated on the centroid of the parabolic curve. The character boards upon the curve are rotated with a slope of the curve under a certain probability. 

\subsubsection{(b) Text Style Selection}

This part chooses colors and textures of a text, and injects additional text effects such as bordering, shadowing and extruding texts.  

A color map $\mathbf{C}$ is an estimation of a real distribution over colors of text images. It can be identified by clustering colors of real text images. It usually consists of 2, or 3 clusters with the mean gray-scale colors and their standard deviation (s.t.d). MJ and ST also utilize this color map identified from ICDAR03 dataset~\cite{IC03} and IIIT dataset~\cite{IIIT5K}, respectively. In our work, we adapt to the color map used in ST. The color selection from the color map is conducted sequentially in an order of a cluster and a color based on the mean and the s.t.d. Once a color is selected, SynthTIGER changes the color of the character appearances.    

The colors of texts in the real world is not simply represented with a single color. SynthTIGER uses multiple texture sources, $\mathbf{B}$, to reflect the realness of text colors. Specifically, it picks up a random texture from $\mathbf{B}$, performs a random crop of the texture, and use it as a texture of the text appearance of the synthetic image. In this process, transparency of the texture is also randomly chosen to diversify the effect of textures.

In the real world, the characters' boundary exhibits diverse patterns depending on text styles, text background, and environmental conditions. We can simulate the boundary styles by applying text border, shadow, and extruding effects. SynthTIGER randomly chooses one of these effects and applies it to the text. All required parameters such as effect size and color will be sampled randomly from a pre-defined range.
 
\subsubsection{(c) Transformation}

The visual appearance of the same scene text image can be significantly different depending on the view angle. Moreover, these text images detected by different OCR engines or labeled by multiple human annotators will have different patterns. SynthTIGER generates synthesized images reflecting these characteristics by utilizing multiple transformation functions.

In detail, SynthTIGER provides stretch, trapezoidate, skew and rotate transformations.
Their functions are explained below.
\begin{itemize}
    \item \textit{Stretch} adjusts the width or height of the text images.
    \item \textit{Trapezoidate} choose an edge of the text image and then adjust its length.  
    \item \textit{Skew} tilts the text image to one of the four directions such as the right, left, top and bottom.  
    \item \textit{Rotate} turns the text image clockwise or anticlockwise.  
\end{itemize}
SynthTIGER applies one of these transformations to the text image with necessary parameter values randomly sampled.

Finally, SynthTIGER adds random margins to simulate the diverse results of text detectors. The margins are independently applied to the top, bottom, left, and right of the image. 

\subsubsection{(d) Blending}

The blending process first creates a background image by randomly sampling color and texture from the color map, $\mathbf{C}$, and the texture database, $\mathbf{B}$. It randomly changes the transparency of the background texture to diversify the impact of the background. Secondly, it creates two text images, foreground and mid-ground, with the same rendering processes but different random parameters. The first one contains a target text and the second one carries a noise text. The next step is to combine the mid-ground and background images. The blending process first crops the background image to match the text image size. Then, it randomly shifts the noise text in the mid-ground and makes the non-textual area transparent. Finally, it merges two images by using one of multiple blending methods: normal, multiply, screen, overlay, hard-light, soft-light, dodge, divide, addition, difference, darken-only, and lighten-only. The last step is to overlay the foreground text image on the merged background. The target text area with a little margin is kept non-transparent to distinguish between the target text and the noise text. During this process, it also uses one of the blending methods aforementioned.

A synthesized image created through these steps from (a) to (d) might not be a good text-focused image for several reasons. For example, its text and background color happen to be indistinguishable because they are chosen independently. To address this problem, we adopted Flood-Fill algorithm\footnote{\href{https://en.wikipedia.org/wiki/Flood\_fill}{https://en.wikipedia.org/wiki/Flood\_fill}}. 
We apply this algorithm starting from a pixel inside the target text, count the number of text boundary pixels visited, and calculate the ratio of the visited text boundary pixels to the number of all boundary pixels. This process is repeated until all target text pixels are used. If this ratio exceeds a certain threshold, we conclude that the target text and background are indistinguishable and discard the generated image.

\subsubsection{(e) Post-processing}

Post-processing is conducted to finalize the synthetic data generation. In this process, SynthTIGER injects general visual noises such as gaussian noise, gaussian blur, resize, median blur and JPEG compression.

\subsection{Text Selection Strategy}

The previous methods, MJ and ST, randomly sample target texts from a user-provided lexicon.
In contrast, SynthTIGER provides two additional strategies to control the text length distribution and character distribution of a synthesized dataset.
It alleviates the long-tail problem inherited from the use of a lexicon.

\subsubsection{Text Length Distribution Control}

The length distribution of texts randomly sampled from a lexicon does not represent the true distribution of a real world text data.
To alleviate this problem, SynthTIGER performs text length distribution augmentation with the probability $p_{\text{l.d.}}$ where $l.d.$ stands for length distribution.
The augmentation process first randomly chooses the target text length between 1 and the pre-defined maximum value.
Then, it randomly samples a word from the lexicon.
If the word matches the target length, SynthTIGER uses it as a target text.
For longer words, it simply cuts off extra rightmost characters.
For shorter words, it samples a new word and attaches it to the right of the previous one until the concatenated word matches or exceeds the target length.
The rightmost extra characters will be cut off.
Text length augmentation, however, should be used with caution because the generated texts are mostly nonsensical.
In the experiment section, we show that text length augmentation with $p_{\text{l.d.}}$=$0.5$ increases STR accuracy more than 2\%, while making little difference with $p_{\text{l.d.}}$=$1.0$.

\subsubsection{Character Distribution Control}

Languages such as Chinese and Japanese use a large number of characters.
A synthesized dataset for such a language often lacks enough amount of samples for rare characters.
To deal with this problem, SynthTIGER conducts character distribution augmentation with the probability $p_{\text{c.d}}$ where $c.d.$ stands for character distribution.
When the augmentation is triggered, it randomly chooses a character from vocabulary and samples a word having that character.
In the experiments, we show that character distribution augmentation with $p_{\text{c.d.}}$ between 0.25 and 0.5 improves STR performance for both scene and document domains.

\section{Experimental Results}

This section consists of (S4.1) our experimental settings for both synthetic data generation and STR model development, (S4.2) comparison to popular synthetic datasets, (S4.3) apples-to-apples comparison of synthesis engines with the common resources, (S4.4) ablative studies on rendering functions, (S4.5) experiments on controlling text distributions. 

\subsection{Experimental Settings}

To compare synthetic data generation engines, synthetic datasets are built with them and then STR performances are evaluated from the models trained with the generated datasets. Here, we describe the resources used in synthetic data generation engines and the training and evaluating settings of STR models. 
\subsubsection{Resources for Synthetic Data Generation}

To build synthetic datasets, multiple resources, $\mathbf{L}$, $\mathbf{F}$, $\mathbf{B}$, and $\mathbf{C}$, are required. Table~\ref{tab:resource-table} describes the resources used in MJ and ST as well as in our experiments. 
As can be seen, MJ and ST are built with their own resources. MJ utilizes a lexicon combining Hunspell corpus\footnote{\href{http://hunspell.github.io/}{http://hunspell.github.io/}} and ground-truths of real STR examples from ICDAR(IC), SVT, and IIIT datasets. MJ also uses textures and its color map of IC03 and SVT. In contrast, ST does not use the ground-truth information except for the color map from IIIT. SynthTIGER utilizes a lexicon consisting of texts of MJ and ST dataset and uses the same textures and color map with ST. They have different number of fonts that are available from google fonts\footnote{\href{https://fonts.google.com/}{https://fonts.google.com/}}. 
Common* in the table uses an another lexicon from Wikipedia to evaluate all synthesis engines without ground-truth information of real STR test examples and test sets except for the color map. For our Japanese STR tasks, we utilize a Japanese lexicon (84M) from Wikipedia and Twitter, 382 fonts, $\mathbf{B}_{\text{ST}}$, and $\mathbf{C}_{\text{ST}}$. 

\begin{table}[t]
\tabcolsep=2.5pt
\centering
\caption{Resources used to build synthetic datasets of Latin texts. Common* indicates a setting for apples-to-apples comparison between synthesis engines. ``$\times$3'' indicates text augmentation for capitalized, upper-cased, and lower-cased words.}
\begin{tabular}{lcccc}
\toprule
& Lexicon ($\mathbf{L}$) & Font ($\mathbf{F}$) & Texture ($\mathbf{B}$) & Color map ($\mathbf{C}$)\\ \midrule
\multirow{2}{*}{MJ~\cite{jaderberg2014synthetic}} & Hunspell + test-sets of & \multirow{2}{*}{1,400 fonts} & IC03, SVT & IC03\\
& IC, SVT, IIIT(90K$\times$3) & & train-set (358) & train-set \\ \midrule
\multirow{2}{*}{ST~\cite{gupta2016synthetic}} & Newsgroup20 & \multirow{2}{*}{1,200 fonts} & Crawling & IIIT \\ 
 & (366K) & & (8,010) & word dataset
\\\midrule
SynthTIGER & MJ + ST (197K$\times$3)  & 3,568 fonts & $\mathbf{B}_{\text{ST}}$ & $\mathbf{C}_{\text{ST}}$  \\ \midrule \midrule
Common* & Wikipedia (19M$\times$3) & $\mathbf{F}_{\text{SynthTIGER}}$ & $\mathbf{B}_{\text{ST}}$ & $\mathbf{C}_{\text{ST}}$ \\ \bottomrule
\end{tabular}
\label{tab:resource-table}
\end{table}

\subsubsection{Experimental Settings for Training and Evaluating STR Models}

In this paper, we evaluate synthetic dataset by training a STR model with them and evaluating the trained model on real STR examples. We choose BEST~\cite{baek2019STRcomparisons} as our base model since it is generally used as well as its implementation is publicly available. All synthetic datasets built for our experiments consists of 10M word box images. The public datasets, MJ and ST, contains 8.9M and 7M word box images respectively and they are also evaluated with the same process. 

The BEST model are trained only with synthetic datasets. The training and evaluation is conducted with the STR test-bed\footnote{\href{https://github.com/clovaai/deep-text-recognition-benchmark}{https://github.com/clovaai/deep-text-recognition-benchmark}}. Most of experimental settings follows the training protocol of Baek et al.~\cite{baek2019STRcomparisons} except for the input image size of 32 by 256.  

The evaluation protocol is also the same with \cite{baek2019STRcomparisons}. Specifically, we test two STR scenario depending on languages: one is Latin and the other is Japanese. For the Latin case, character vocabulary consists of 94 including both alphanumeric and special characters. 
STR models are evaluated on test-sets of STR benchmarks; 3,000 images of IIIT5k~\cite{IIIT5K}, 647 images of SVT~\cite{SVT}, 1,110 images of IC03~\cite{IC03}, 1,095 images of IC13~\cite{IC13}, 2,077 images of IC15~\cite{IC15}, 645 images of SVTP~\cite{SVTP}, and 288 images of CUTE80~\cite{CUTE80}. We also test performances on business documents with our in-house 38,493 images. 
We only evaluate on alphabets and digits due to in-consistent labels of the benchmark datasets. For the Japanese case, the vocabulary consists of 6,723 characters including alphanumeric, special, hiragana/katakana, and some Chinese characters. The evaluation is conducted on our in-house datasets; 40,938 images of scenes and 38,059 images of Japanese business documents.

\subsection{Comparison on Synthetic Text Data}

\begin{table}[t!]
\tabcolsep=4pt
\centering
\caption{Benchmark performances of BEST~\cite{baek2019STRcomparisons} trained from synthetic text images. The amount of MJ, ST, MJ+ST, our data are 8.9M, 7M, 15.9M, and 10M, respectively.}
\begin{tabular}{lcccccccc}
\toprule
\multicolumn{1}{c}{\multirow{2}{*}{Dataset}} & \multicolumn{4}{c}{Regular} & \multicolumn{3}{c}{Irregular} & \multicolumn{1}{c}{\multirow{2}{*}{Total}} \\ \cmidrule(lr){2-5} \cmidrule(lr){6-8}
& IIIT5k & SVT & IC03 & IC13 & IC15 & SVTP & CUTE80 & \\ \midrule
MJ~\cite{jaderberg2014synthetic} & 83.4 & 84.5 & 85.6 & 83.5 & 66.0 & 73.0 & 64.6 & 78.3 \\
ST~\cite{gupta2016synthetic} & 86.1 & 82.5 & 90.7 & 89.8 & 64.5 & 69.1 & 60.1 & 79.7 \\
MJ~\cite{jaderberg2014synthetic} + ST~\cite{gupta2016synthetic} & 90.9 & 87.2 & \textbf{92.1} & 91.2 & \textbf{72.9} & \textbf{77.8} & 73.6 & 85.1 \\
SynthTIGER (Ours) & \textbf{93.2} & \textbf{87.3} & 90.5 & \textbf{92.9} & 72.1 & 77.7 & \textbf{80.6} & \textbf{85.9} \\ \bottomrule
\end{tabular}
\label{tab:main-table}
\end{table}

Table~\ref{tab:main-table} compares STR performances of BEST~\cite{baek2019STRcomparisons} models trained with synthetic text images from MJ, ST, and SynthTIGER. 
As reported in previous works, the combination of MJ and ST shows better performances than their single usages. SynthTIGER always provides a better STR performance than the single usages of MJ and ST. Interestingly, ours achieves comparable or better performance than combined data (MJ+ST). It should be noted that the amount of combined training data is 1.5 times larger than ours.

\begin{table}[t!]
\tabcolsep=4pt
\centering
\caption{Benchmark performances of BEST~\cite{baek2019STRcomparisons} trained from synthetic generators with \textit{the same resources}. The total amount of MJ*, ST*, MJ*+ST*, and Ours* are identical to 10M.}
\begin{tabular}{lcccccccc}
\toprule
\multicolumn{1}{c}{\multirow{2}{*}{Dataset}} & \multicolumn{4}{c}{Regular} & \multicolumn{3}{c}{Irregular} & \multicolumn{1}{c}{\multirow{2}{*}{Total}} \\ \cmidrule(lr){2-5} \cmidrule(lr){6-8}
& IIIT5k & SVT & IC03 & IC13 & IC15 & SVTP & CUTE80 & \\ \midrule
MJ* & 87.1 & 81.9 & 83.6 & 86.1 & 62.9 & 69.8 & 57.3 & 78.3 \\
ST* & 79.0 & 80.4 & 76.8 & 79.5 & 59.6 & 66.2 & 59.4 & 72.8 \\
MJ* + ST* & 89.5 & 83.9 & \textbf{87.0} & \textbf{89.3} & 68.1 & \textbf{74.6} & 72.9 & \textbf{82.1} \\
SynthTIGER* (Ours*) & \textbf{89.8} & \textbf{84.5} & 84.2 & 87.9 & \textbf{69.5} & 73.8 & \textbf{74.0} & \textbf{82.1} \\ 
\bottomrule
\end{tabular}
\label{tab:main-table-star}
\end{table}

\subsection{Comparison on Synthetic Text Image Generators with Same Resources}

Since the outputs of the engines depend on the resources such as fonts, textures, color maps, and a lexicon, we provide fair comparisons, referred as to `*', by setting the same resources in Table~\ref{tab:main-table-star}.
To present a fair comparison, we set the total amount of comparison data as 10M. For example, the total amount of MJ*+ST* is 10M and other comparisons such as MJ*, ST*, and Ours* are also 10M.
In Table~\ref{tab:main-table-star}, ours shows clear improvement from single usages of MJ* and ST*. Also, ours have comparable performance with combined datasets.

\begin{figure}[t]
\centering
\includegraphics[width=1.0\linewidth]{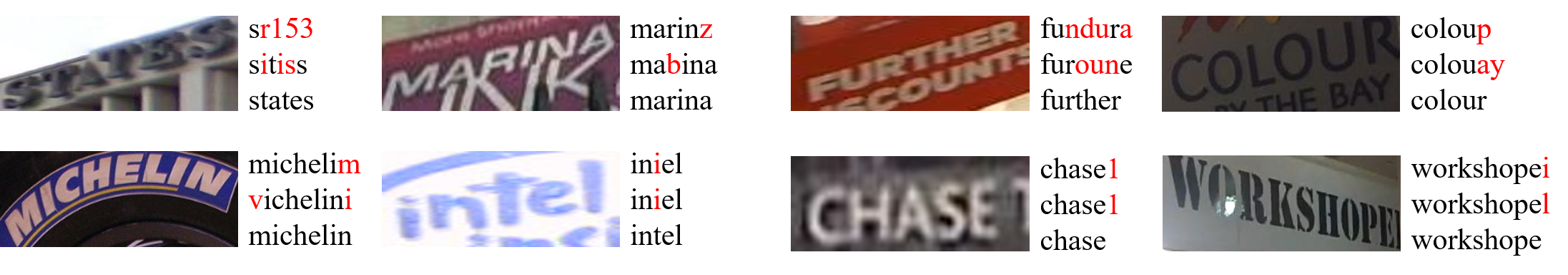}
\caption{Correctly predicted cases of ours. 
The predictions are positioned on the right side of the images.
The texts of the first row are the predictions of MJ*+ST*; The texts of the second row are the predictions of Ours* (-) Mid-ground text that is described in ablative study;    
The texts of the third row are the predictions of Ours*.}
\label{fig:prediction-results}
\end{figure}

We collect some examples from the test benchmarks where ours provides correct predictions. In Fig.~\ref{fig:prediction-results}, ours has robust predictions when the text-images contain high-frequency noises such as lines, complex backgrounds, and parts of other characters. 
Although complexly combined functions could contribute to the correct predictions, 
we believe employing the proposed mid-ground also result in robust performance. 
We will provide more descriptions about effects of mid-ground in (S4.4) with the ablative study.

\begin{table}[t!]
\tabcolsep=4pt
\centering
\caption{Latin and Japanese recognition performance on scene and document images.}
\begin{tabular}{lcccc}
\toprule
\multicolumn{1}{c}{\multirow{2}{*}{Dataset}} & \multicolumn{2}{c}{Latin} & \multicolumn{2}{c}{Japanese} \\ \cmidrule(lr){2-3} \cmidrule(lr){4-5}
& Scene & Document & Scene & Document \\ \midrule
MJ* + ST* & \textbf{82.1} & 82.5 & 60.0 & 83.8 \\
Ours* & \textbf{82.1} & \textbf{85.6} & \textbf{60.1} & \textbf{86.8} \\ \bottomrule
\end{tabular}
\label{tab:sub-table-lang}
\end{table}

We extend the experiments on Japanese to compare the language generalization performance between ours and previous generators.
Moreover, we demonstrate the performance of document images to confirm the extensibility of another domain.
In these experiments, all engines share the same resources because MJ and ST have not provided Japanese.
Table~\ref{tab:sub-table-lang} shows that ours achieves comparable or better performance than combined datasets. Specifically, ours achieves much better performance on document image with 3.0 pp improvements. 
Since document images usually contain high-frequency noises such as scan noise and part of characters that are included in other lines or paragraphs, we believe the use of mid-ground could cope with these noises.

\subsection{Ablative Studies on Rendering Functions}\label{sec:ablative}
\begin{figure}[t]
\centering
\includegraphics[width=1.0\linewidth]{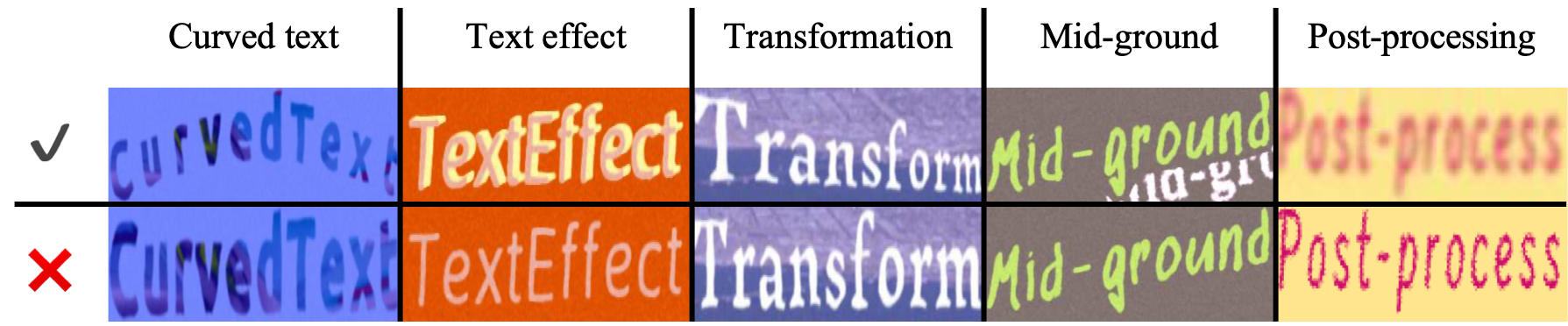}
\caption{Rendering effect visualization of SynthTIGER modules. The images on the top show the cases when the effects are applied. The bottom images represent the cases when the effects are off. All functions are essential to reveal the realness of the synthetic text images.}
\label{fig:rendering-functions}
\end{figure}

We investigate the effects on the STR performance of rendering function by excluding each function.
Although, we do not propose and optimize all the rendering functions, 
we believe these ablative studies are significant to STR fields.
This is because detailed investigations of rendering functions have not been reported and also can be a guide to subsequent data generation researches.   
To help intuitive understanding of each function, we present some visual examples of rendering function in Fig.~\ref{fig:rendering-functions}.
As presented in Table~\ref{tab:rendering-table}, the texture blending, transformation, margin, and post-processing critically impact the performance. 
Moreover, the proposed mid-ground text enhances both regular and irregular benchmarks.
Other rendering functions also contribute to the STR performances.

To show the effects of the proposed mid-ground, we present some examples in Fig.~\ref{fig:prediction-results} where baseline (Ours*) can correctly predict the results, but ``(-)Mid-ground text'' cannot. These figures show that the proposed mid-ground can help to handle more diverse real-world scenes that could degrade the recognition performances. 

\begin{table}[t]
\tabcolsep=8pt
\centering
\caption{The performance of rendering functions. (-) indicates exclusion from baseline. The color exclusion indicates using random color selection and the blending exclusion means using ``normal'' blending}
\begin{tabular}{lccc}
\toprule
& Regular & Irregular & Total \\ \midrule
Baseline & 87.8 & 70.9 & 82.1 \\ \midrule
(-) Curved text & 87.2 (-0.6) & 70.2 (-0.7) & 81.4 (-0.7) \\
(-) Elastic distortion & 87.7 (-0.1) & 68.4 (-2.5) & 81.1 (-1.0) \\
(-) Color map & 87.6 (-0.2) & 70.7 (-0.2) & 81.9 (-0.2) \\
(-) Texture blending & 84.5 (-3.3) & 66.0 (-4.9) & 78.2 (-3.9) \\
(-) Text effect & 87.3 (-0.5) & 67.7 (-3.2) & 80.6 (-1.5) \\
(-) Transformation & 87.3 (-0.5) & 64.4 (-6.5) & 79.5 (-2.6) \\
(-) Margin & 87.1 (-0.7) & 66.7 (-4.2) & 80.2 (-1.9) \\
(-) Mid-ground text & 87.4 (-0.4) & 69.6 (-1.3) & 81.4 (-0.7) \\
(-) Blending modes & 88.0 (+0.2) & 70.2 (-0.7) & 81.9 (-0.2) \\
(-) Visibility check & 87.1 (-0.7) & 71.3 (+0.4) & 81.8 (-0.3) \\
(-) Post-processing & 86.1 (-1.7) & 59.0 (-11.9) & 76.9 (-5.2) \\ \bottomrule
\end{tabular}
\label{tab:rendering-table}
\end{table}

\begin{figure}[t]
\centering
\subfigure[Length distributions]{\label{fig:length-distribution}\includegraphics[width=0.49\linewidth]{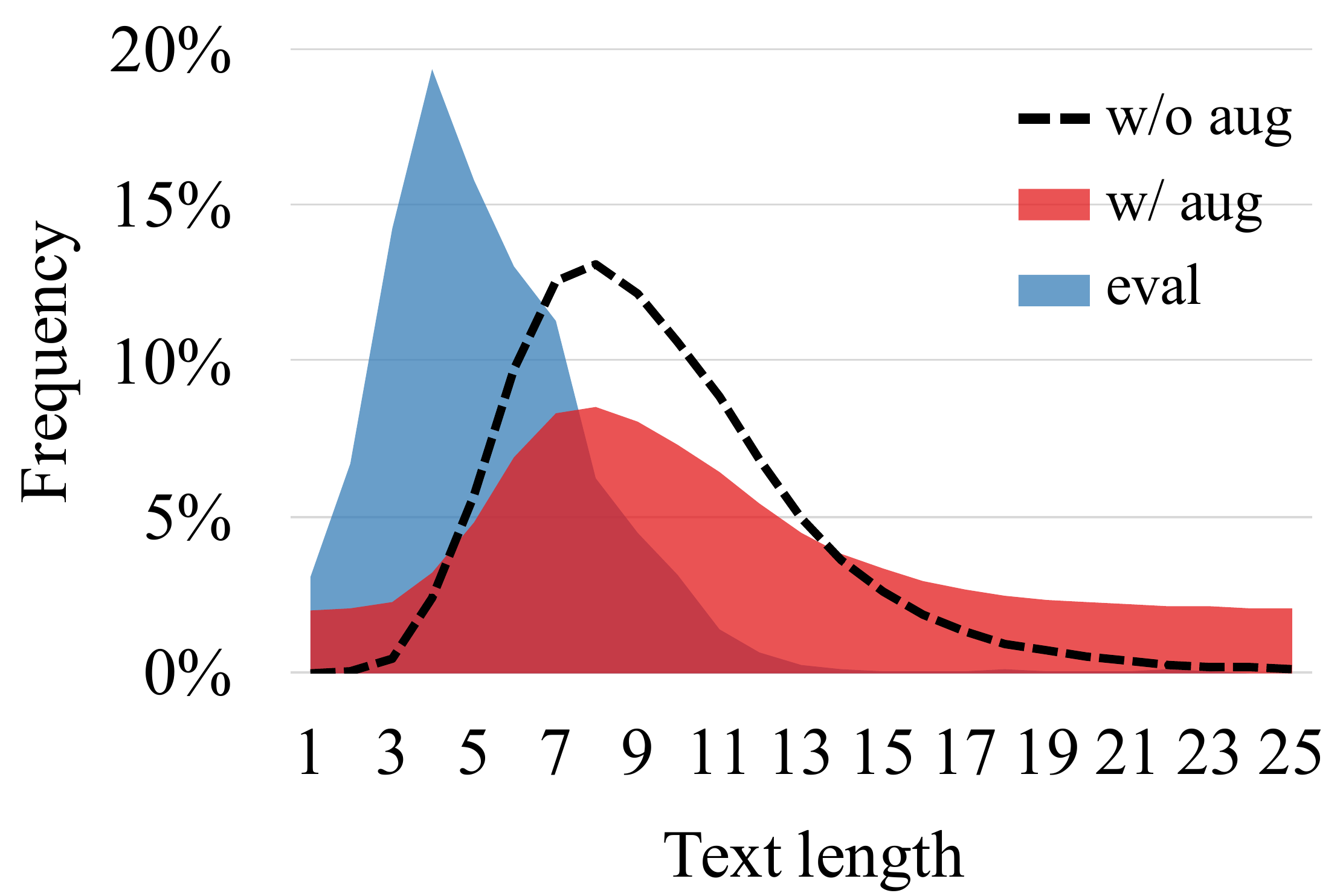}}
\subfigure[STR performances]{\label{fig:length-accuracy}\includegraphics[width=0.49\linewidth]{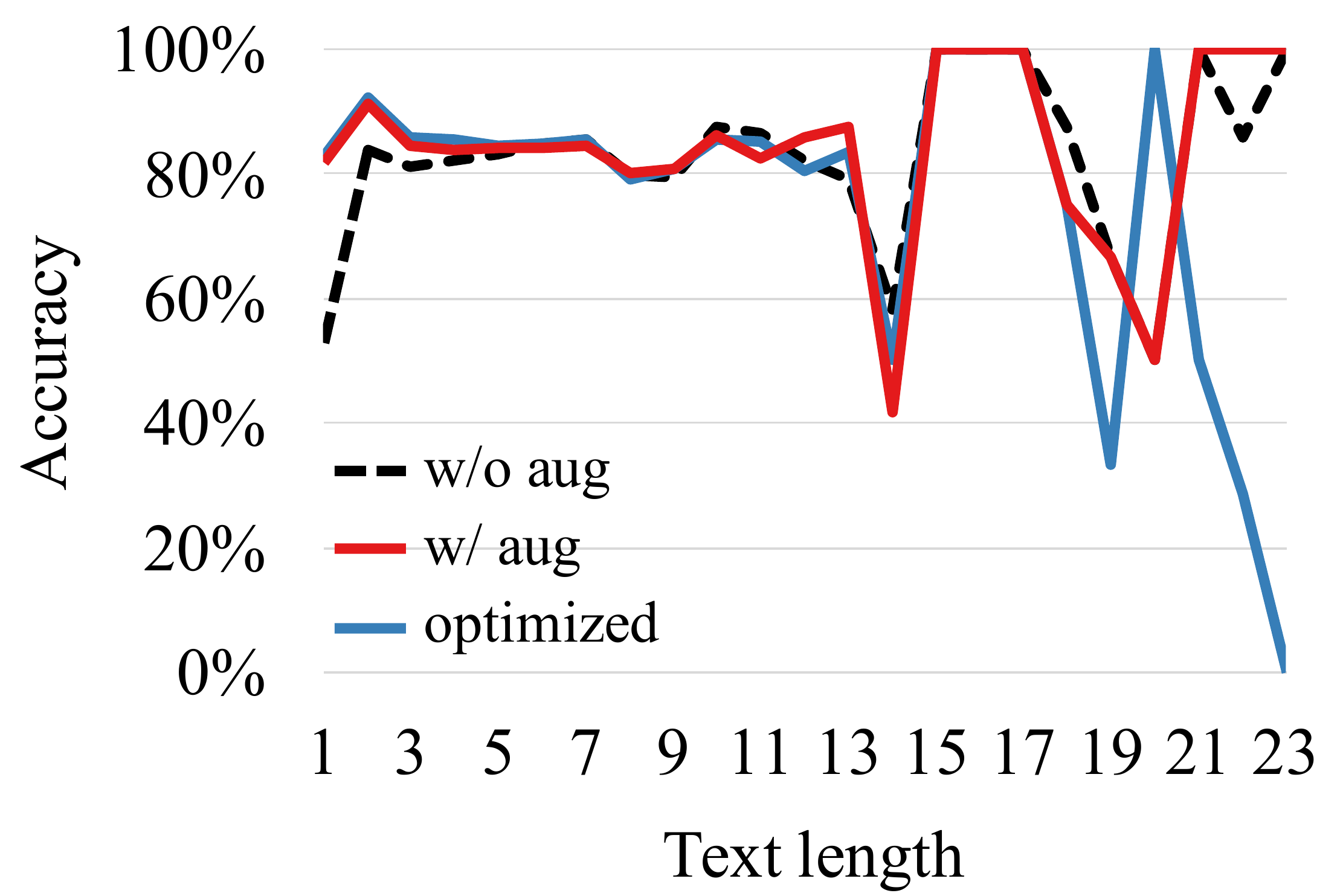}}
\caption{(a) Text length distribution of training data without augmentation (dashed), with augmentation (red) and evaluation data (blue). (b) Accuracy by the length of the model without length augmentation (dashed), model with 50\% applied (red) and model with length augmentation optimized to length distribution of evaluation data (blue). The blue line drops sharply for long texts because texts longer than 15 characters in the evaluation dataset rarely exist. }
\label{fig:length-control}
\end{figure}

\begin{table}[t]
\tabcolsep=8pt
\centering
\caption{STR accuracy according to the change of probability of length distribution augmentation.``Optimized'' indicates the length augmentation is optimized to length distribution of evaluation data.}
\begin{tabular}{cccccc|c}
\toprule
Probability & 0\% & 25\% & 50\% & 75\% & 100\% & Optimized \\ \midrule
Accuracy & 82.1 & 83.9 & \textbf{84.2} & 82.7 & 82.0 & 84.9 \\ \bottomrule
\end{tabular}
\label{tab:length-table}
\end{table}

\subsection{Experiments on Text Selection}

\subsubsection{Experiments on Text Length Distribution Control}
As presented in Fig.~\ref{fig:length-distribution}, we found that 
short and long length texts in Latin training data are insufficient to cover real-world texts and 
length distribution between training and evaluation data are quite different.
To alleviate these problems, we apply text length distribution augmentation to Latin, and thus, the augmented distribution can cover a wide range of length texts as a red graph.
We also present the STR performance according to the change of $p_{\text{l.d.}}$ in Table~\ref{tab:length-table}.
We find that the best performance is achieved when $p_{\text{l.d.}}$ sets 50\% and it is comparable with ``Optimized'' that is regarded as upper bound. 
Specifically, ``Optimized'' makes the distribution of training data have a similar distribution of evaluation data by controlling target length.
Fig.~\ref{fig:length-accuracy} shows that the proposed augmentation prevents critical performance degradation for very short and long length texts. 

\subsubsection{Experiments on Character Distribution Control}
\begin{figure}[t]
\centering
\subfigure[Character distributions]{\label{fig:char-distribution}\includegraphics[width=0.49\linewidth]{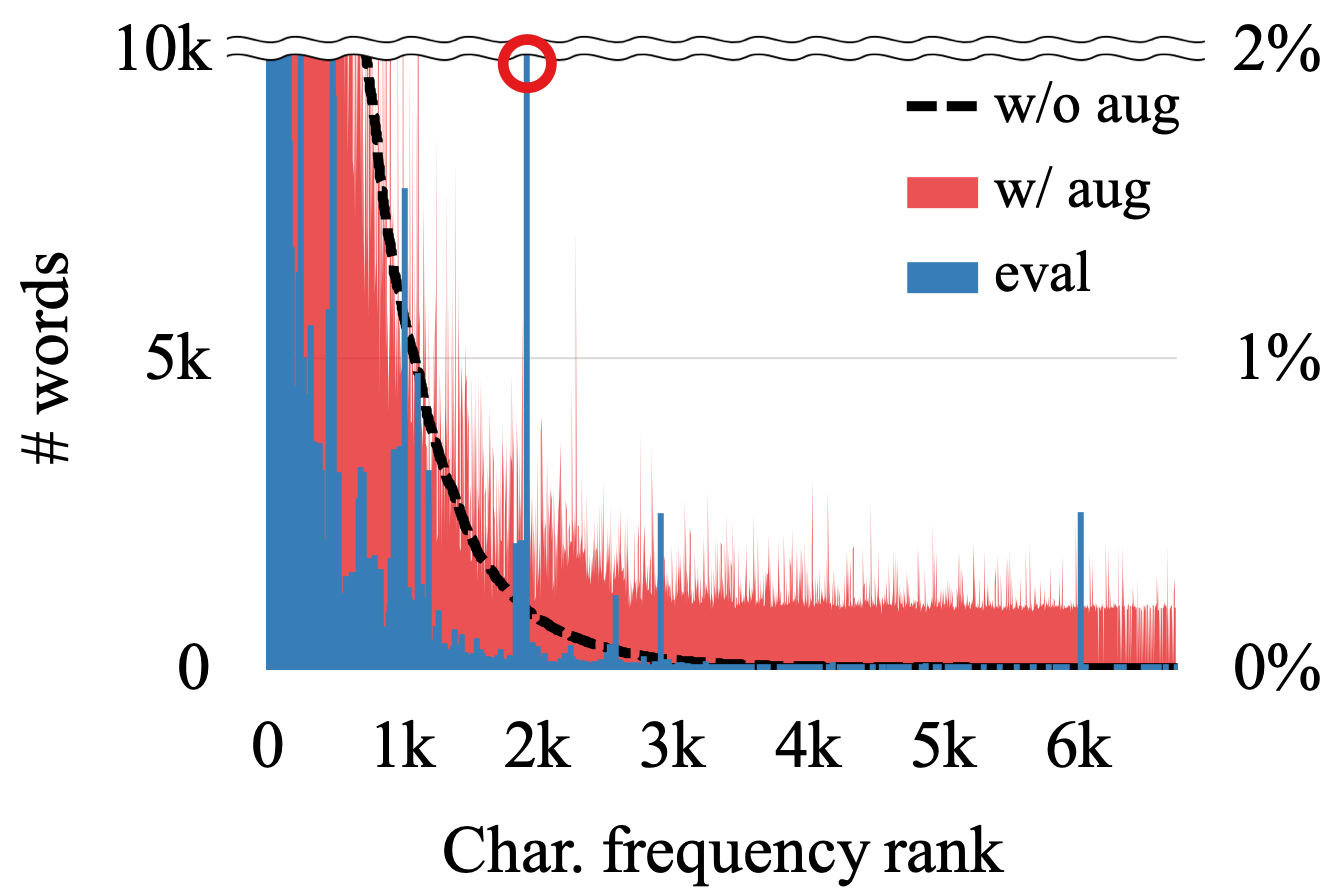}}
\subfigure[STR performances]{\label{fig:char-accuracy}\includegraphics[width=0.49\linewidth]{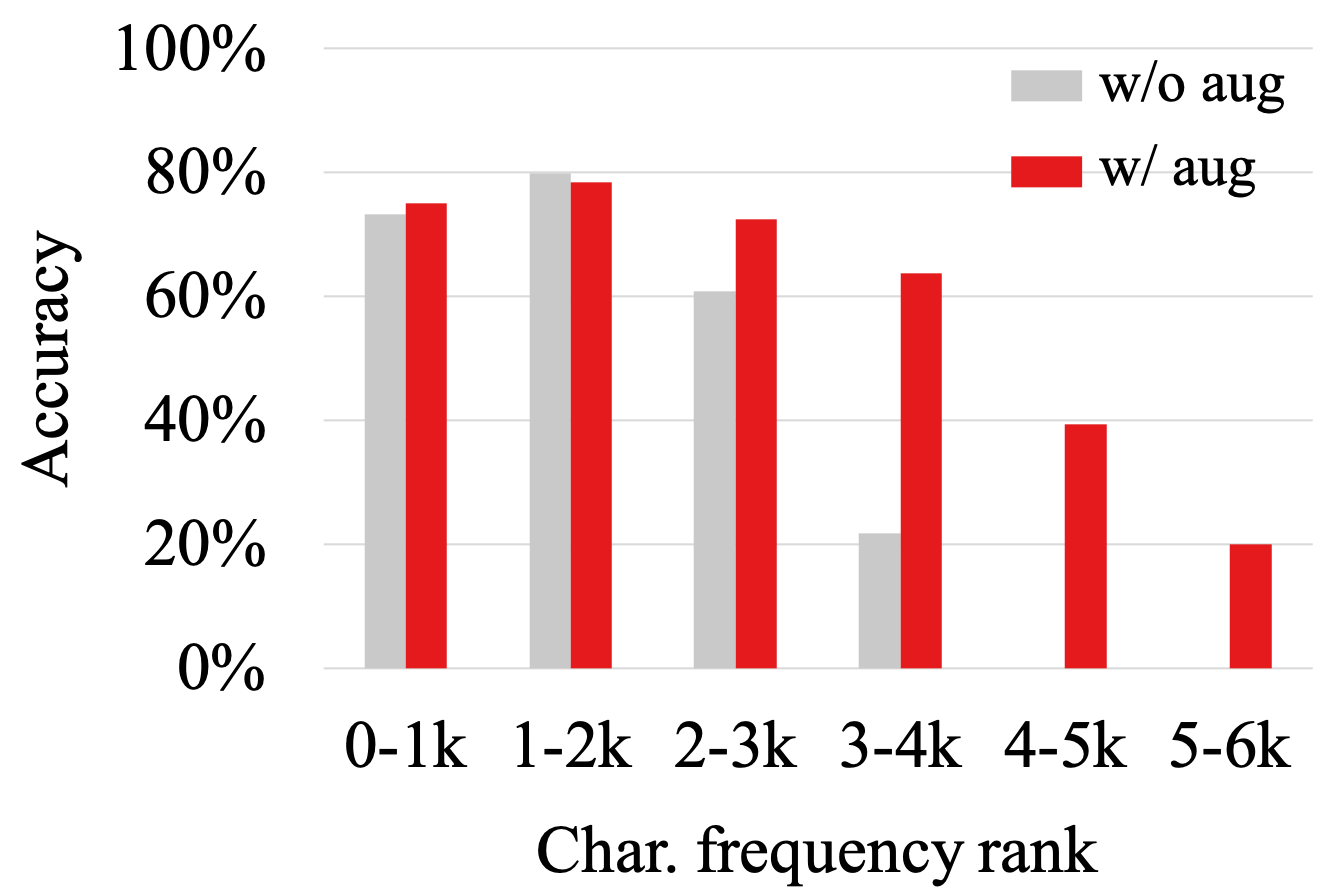}}
\caption{(a) The black dashed line shows the relationship between a specific Japanese character (X-axis) and the number of words involving that character (Y-axis) in the training data. Note that characters at the X-axis are sorted in descending order by the number of their occurrence. The red and blue histograms show the same relationship with the character distribution augmented training data and the evaluation data. (b) The red and gray histograms indicate the STR accuracy (Y-axis) for a subset of the text vocabulary (X-axis). Note that the values at X-axis (e.g., 0-1k) stands for a set of any words involving a character that occurs $min$-$max$ times in the training data.}
\label{fig:char-control}
\end{figure}

As presented in Fig.~\ref{fig:char-distribution}, Japanese characters, which is composed of thousands of characters, have an unbalanced long-tail problem. However, these low-presented characters in training data are frequently exhibited in evaluation data. 
For example, ¥, which is a currency sign with red-circled in the figure, contained words are rarely presented in training data, but, they are frequently exhibited in evaluation data.
To relieve this problem, we apply character augmentation method guaranteeing the minimum numbers of examples including rare characters as red histogram in Fig.~\ref{fig:char-distribution}.
We also present text recognition performances according to the change of $p_{\text{c.d.}}$ in Table~\ref{tab:char-table}. It shows the character distribution augmentation greatly improves scene and document performances when the $p_{\text{c.d.}}$ sets 50\%. It can be seen in Fig.~\ref{fig:char-accuracy} that the proposed augmentation works for recognizing rare character included words.

\begin{table}[t]
\tabcolsep=8pt
\centering
\caption{Recognition accuracy according to the change of probability of character distribution augmentation.}
\begin{tabular}{cccccc}
\toprule
Probability & 0\% & 25\% & 50\% & 75\% & 100\% \\ \midrule
Scene & 60.1 & 61.9 & \textbf{62.6} & 59.4 & 54.5 \\
Document & 86.8 & \textbf{88.1} & 87.8 & 87.2 & 83.2 \\ \bottomrule
\end{tabular}
\label{tab:char-table}
\end{table}

\section{Conclusion}

Synthesizing text images is essential to learn a general STR model by simulating diversity of texts in the real-world. However, there has been no guideline for the synthesis process. This paper addresses the issues by introducing a new synthesis engine, SynthTIGER, for STR. SynthTIGER solely shows better or comparable performance when compared to existing synthetic datasets and its rendering functions are evaluated under a fair comparison. SynthTIGER also addresses biases on text distribution of synthetic datasets by providing two text selection methods over lengths and characters. Our experiments on rendering methods and text distributions show that controlling text styles and text distributions of synthetic dataset affects to learn more generalizable STR models. Finally, this paper contributes to OCR community by providing an open-sourced synthesis engine and a new synthetic dataset.

\bibliography{reference}
\bibliographystyle{splncs04}

\end{document}